\documentclass[pdflatex,sn-vancouver-num]{sn-jnl}

\usepackage{graphicx}%
\usepackage{multirow}%
\usepackage{amsmath,amssymb,amsfonts}%
\usepackage{amsthm}%
\usepackage{mathrsfs}%
\usepackage[title]{appendix}%
\usepackage{xcolor}%
\usepackage{textcomp}%
\usepackage{manyfoot}%
\usepackage{booktabs}%
\usepackage{algorithm}%
\usepackage{algorithmicx}%
\usepackage{algpseudocode}%
\usepackage{listings}%
\usepackage{makecell}
\usepackage{pifont}
\usepackage{geometry}
\usepackage{lmodern}
\usepackage{enumitem}
\usepackage{verbatim}
\usepackage[most]{tcolorbox}
\usepackage{caption}
\usepackage[export]{adjustbox}

\usepackage{booktabs}
\usepackage{tabularx}
\usepackage{caption}

\usepackage{tcolorbox}
\usepackage{listings}
\usepackage{enumitem}

\lstdefinestyle{jsonstyle}{
  basicstyle=\ttfamily\small,
  breaklines=true,
  columns=fullflexible
}

\lstdefinestyle{tabstyle}{
  basicstyle=\ttfamily\small,
  breaklines=true,
  columns=fullflexible
}

\usepackage{listings}
\theoremstyle{thmstyleone}%
\theoremstyle{thmstyletwo}%

\theoremstyle{thmstylethree}%

\begin{document}

\title[Article Title]{TableZoomer: A Collaborative Agent Framework for Large-scale Table Question Answering}


\author[]{\fnm{Sishi} \sur{Xiong}}\email{xiongsishi@chinatelecom.cn}
\equalcont{These authors contributed equally to this work.}
\author[]{\fnm{Ziyang} \sur{He}}\email{e1553228@u.nus.edu}
\equalcont{These authors contributed equally to this work.}
\author[]{\fnm{Zhongjiang} \sur{He}}\email{hezj@chinatelecom.cn}
\author[]{\fnm{Yu} \sur{Zhao}}\email{zhaoyu11@chinatelecom.cn}
\author[]{\fnm{Changzai} \sur{Pan}}\email{panzz3@chinatelecom.cn}
\author[]{\fnm{Jie} \sur{Zhang}}\email{zhangj157@chinatelecom.cn}
\author[]{\fnm{Zhenhe} \sur{Wu}}\email{wuzhenhe@buaa.edu.cn}
\author*[ ]{\fnm{Shuangyong} \sur{Song}}\email{songshy@chinatelecom.cn}
\author*[ ]{\fnm{Yongxiang} \sur{Li}}\email{liyx25@chinatelecom.cn}

\affil[]{\orgdiv{Institute of Artificial Intelligence (TeleAI)}, \orgname{China Telecom Corp Ltd}, \city{Beijing}, \country{China}}

\abstract{While large language models (LLMs) have shown promise in the table question answering (TQA) task through prompt engineering, they face challenges in industrial applications, including structural heterogeneity, difficulties in target data localization, and bottlenecks in complex reasoning. To address these limitations, this paper presents TableZoomer, a novel LLM-powered, programming-based agent framework. It introduces three key innovations: (1) replacing the original fully verbalized table with structured table schema to bridge the semantic gap and reduce computational complexity; (2) a query-aware table zooming mechanism that dynamically generates sub-table schema through column selection and entity linking, significantly improving target localization efficiency; and (3) a Program-of-Thoughts (PoT) strategy that transforms queries into executable code to mitigate numerical hallucination. Additionally, we integrate the reasoning workflow with the ReAct paradigm to enable iterative reasoning. Extensive experiments demonstrate that our framework maintains the usability advantages while substantially enhancing performance and scalability across tables of varying scales. When implemented with the Qwen3-8B-Instruct LLM, TableZoomer achieves accuracy improvements of 19.34\% and 25\% over conventional PoT methods on the large-scale DataBench dataset and the small-scale Fact Checking task of TableBench dataset, respectively.\renewcommand{\thefootnote}{\arabic{footnote}}\footnote[1]{Our codes are available at \url{https://github.com/ccx06/TableZoomer}.}}


\maketitle

\section{Introduction}\label{sec1}

Table Question Answering, a prominent subtask within the broader field of question answering (QA), focuses on extracting or inferring answers to natural language questions from tabular data \citep{jin2022surveytablequestionanswering}. In the data-driven era, tables have emerged as a core data medium widely adopted across diverse domains due to their efficiency in storing massive volumes of structured information. Consequently, table understanding tasks have attracted growing research interest, with TQA constituting a critical component of these tasks.

Although LLMs demonstrate remarkable capabilities in natural language processing and can be adapted to diverse downstream tasks via prompt engineering, their inherent text-based retrieval and reasoning mechanisms face significant limitations when handling large-scale, heterogeneous tabular data~\citep{ruan2024languagemodelingtabulardata}. Compared to free-text QA, TQA poses distinct challenges, primarily requiring models to perform multi-step precise reasoning under structured data constraints. Key challenges include: (i) Structural heterogeneity. Tabular data often contains heterogeneous features (e.g., numerical, categorical, textual) with implicit inter-row/column relationships, complicating unified representation learning and structured semantic alignment. (ii) Challenges in target data localization. For a single QA instance, most data in the table is irrelevant, introducing noise interference \citep{huang2025learnbeneficialnoisegraph, 10.1609/aaai.v39i16.33918, zhang2024dataaugmentationcontrastivelearning, zhang2025variationalpositiveincentivenoisenoise, Li_2024}. Large-scale tables are also susceptible to truncation due to input sequence length constraints. Concurrently, when processing lengthy table sequences, models exhibit degradation in the quality of vector representations, and intermediate processing stages often lead to information loss. These factors substantially impede localization of key information and feature extraction. (iii) Complex reasoning bottlenecks: Textual Chain-of-Thought (TCoT) approaches exhibit limitations in logical inference (e.g., cross-cell calculations, comparisons) and numerical operations, frequently leading to factual errors.

To unlock the potential of LLMs in table understanding, recent research has focused on two main optimization strategies. The first strategy involves specialized pre-training or post-training of LLMs specifically for tables, exemplified by models like TableLlaMa \citep{zhang-etal-2024-tablellama}, TableGPT2 \citep{su2024tablegpt2largemultimodalmodel}, StructLM \citep{zhuang2024structlm}, TableLLM \citep{zhang2025tablellmenablingtabulardata} and the reinforcement learning-based Table-R1 
\citep{wu2025tabler1regionbasedreinforcementlearning}. However, these approaches typically demand extensive domain-specific tabular data and entail high computational costs. 
The second strategy leverages LLM-driven agent workflow. Complex problems are decomposed and solved incrementally through predefined planning. Some techniques within this paradigm first extract relevant sub-tables from the original table to reduce context size, as employed in methods such as Chain-of-Table \citep{wang2024chainoftable}, Dater \citep{dater}, and TableMaster \citep{cao2025tablemasterrecipeadvancetable}.
A key limitation of these methods is the initial requirement to utilize the entire table (often formatted as Markdown or JSON) as context, which significantly increases computational complexity, token consumption, and risks information loss. Furthermore, generating and then executing SQL or Python code to enhance numerical computation has emerged as a popular and effective approach within agent workflows, implemented in framework like Binder \citep{cheng2023binding}, LEVER \citep{ni2023lever}, and TableRAG \citep{chen2024tablerag}.

Despite these advances, the current research on TQA generally overlooks the practical challenges posed by large-scale tables in real-world industrial scenarios. The limited context window size of LLMs and Loss in the Middle \citep{liu-etal-2024-lost} phenomenon make processing large-scale tables difficult or even infeasible. Crucially, there is a lack of fundamental TQA framework designed for large-scale tables.
To address this gap, we propose TableZoomer, an effective LLM-driven agent framework. It aims to efficiently and accurately handle information localization and QA, particularly for large-scale tables. Our key innovations comprise:

\textbf{(1) Schema-based Complexity Reduction}. Inspired by Text-to-SQL approaches~\citep{wang-etal-2025-mac, kobayashi-etal-2025-read, 10096172}, we extract and utilize the schema mode to capture global information of tables. This reduces the inherent data complexity from $O(M\times N)$ to $O(N)$, where $M$ and $N$ represent the number of rows and columns respectively. Semantic and statistical features are integrated into the schema representation to enhance table understanding.

\textbf{(2) Query-aware Zooming and Focusing}. Recognizing that most data in the table is query-specific redundant, we propose a query-aware zooming mechanism to dynamically derive a minimal query-relevant sub-schema from the global table schema, enabling LLM to understand the table from a more focused perspective. This mechanism employs column-level filtering via column selection and row-level localization combined with entity linking. It significantly reduces interference from extraneous information while further minimizing token consumption.

\textbf{(3) Robust Execution and Interactive Reasoning}. To improve the accuracy of numerical computations and mitigate hallucinations, we adopt the Program-of-Thoughts (PoT) \citep{chen2023programthoughtspromptingdisentangling} reasoning method. This guides the LLM to generate executable Python code for deriving results, incorporating a built-in error feedback mechanism to enhance robustness. Furthermore, we integrate the table reasoning workflow into an iterative thinking paradigm ReAct \citep{yao2023react}, allowing incremental cycles of thinking, reasoning and reflection. 

TableZoomer significantly enhances QA performance on tabular data while maintaining low computational overhead, systematically addressing key bottlenecks in LLM-based large-scale table processing. It is flexible to any advanced language model, and delivers excellent results even without fine-tuning. Our main contributions are:

\begin{itemize}
    \item We propose a novel LLM-powered agent framework optimized for large-scale table question answering that delivers robust performance with significantly lower token consumption.\\
    \item We perform extensive experiments on three popular datasets with varying table sizes, demonstrating the proposed framework's effectiveness, efficiency, and scalability. \\
    \item We further conduct multi-dimensional and error analyses to thoroughly demonstrate its inner workings.
\end{itemize}

\section{Related Works}\label{sec2}

TQA tasks fundamentally involve interpreting and manipulating structured tabular data to answer specific questions, verify facts, or generate concise summaries. Early works primarily interact with tabular data through executable programming language queries \citep{yin2016neuralenquirerlearningquery, DBLP:conf/semweb/LiWLHFZZLLS24, wu-etal-2025-ucs, MR-SQL} or by leveraging graph neural networks and convolutional neural networks \citep{zhong-etal-2020-logicalfactchecker, yu2019spiderlargescalehumanlabeleddataset} to better encode and capture the internal structure of tables \citep{wu2024rbsqlretrievalbasedllmframework}. However, these methods often focus on the local or structural features of tables, showing clear limitations when tackling tasks that demand complex logical reasoning across multiple rows and columns, deep semantic understanding, or the capture of abstract concepts.

To support reasoning over tabular data with LLMs, researchers have developed a wide range of techniques. These approaches can be broadly classified into three methodological categories: prompt-based techniques, which rely on carefully designed prompts to elicit reasoning; domain training, which adapts models using annotated table corpora; and LLM-based agent frameworks, which decompose the task into multiple tool-augmented steps. We discuss each category in detail below.

\subsection{Prompt‑based Techniques}
LLMs excel at in-context learning \citep{wang-etal-2024-sentence}, where a few demonstrations are sufficient to guide reasoning without updating model parameters. The Chain-of-Thought (CoT) prompt-based methods \citep{wei2023chainofthoughtpromptingelicitsreasoning,zhang2024lemur} introduce explicit intermediate reasoning steps, paving the way for techniques such as Least-to-Most Prompting (LtM) \citep{zhou2023leasttomostpromptingenablescomplex}, which guides models through complex tasks in a structured manner. Among these methods, TCoT generates intermediate reasoning steps in natural language to facilitate multi-step logical inference. In contrast, PoT converts the original problem into executable code (e.g., Python) and obtains precise results through program execution—demonstrating strong performance in tasks requiring numerical computation or structured data deduction.

Despite these strengths, prompt-based approaches face critical limitations when applied to TQA. First, the inherently two-dimensional structure of tabular data fundamentally differs from linear text, making it difficult for prompting strategies to capture fine-grained row-column relationships. In scenarios requiring cross-row or hierarchical comparisons, LLMs often produce faulty reasoning chains with incorrect cell references or semantic mismatches. Second, a major bottleneck lies in the trade-off between table size and the LLM’s context length. Feeding a full table with million of cells can exhaust the token budget and leave little space for reasoning steps, while feeding only partial content risks omitting key information. This token-context dilemma severely restricts the applicability of prompt-based methods in real-world table reasoning tasks.

\subsection{Domain Training}
To better model the structural and semantic complexity of tabular data, a line of work adopts pre-training on  domain-specific corpora \citep{LIU2023127, Liu2024icsPLMsEP, Song2020SentimentAT}, aiming to equip language models with table-aware inductive biases. These models capture row-column alignment patterns, schema semantics, and numerical regularities during pre-training. Building upon such initialization, recent studies further fine-tune decoder-only models \citep{touvron2023llamaopenefficientfoundation}—such as LLaMA or GPT —with supervised objectives tailored to table question answering and reasoning. Notable table LLMs include TableLLaMa, TableGPT2, StructLLM \citep{zhuang2024structlmbuildinggeneralistmodels}, and TableLLM, which leverage synthetic or annotated data for schema linking, cell-level inference, and numerical reasoning. Compared to prompt-based methods, fine-tuning approaches often yield more stable and accurate performance, especially when task-specific supervision is available. They are particularly effective at internalizing symbolic patterns and aligning tabular semantics in a data-efficient manner during inference.

In real-world deployments, however, fine-tuned models often struggle to generalize and scale. First, they typically require large quantities of high-quality labeled table data, which is expensive and time-consuming to collect across different domains. Moreover, once fine-tuned, these models lack flexibility for task composition or adaptation to unseen table formats, and retraining incurs high costs when task distributions shift. Additionally, the pre-training corpora used for table modeling are often synthetic or domain-biased, limiting generalization.These challenges have motivated growing interest in alternative paradigms.

\subsection{LLM-based Agent Framework}
To address the data annotation bottleneck faced by domain fine-tuning approaches, recent research has increasingly shifted toward LLM-based agent workflows. These frameworks coordinate multiple specialized modules to jointly accomplish table reasoning tasks, offering greater flexibility and reducing reliance on supervised data. Representative works include TableRAG, TableMaster, OpenTab \citep{kong2024opentabadvancinglargelanguage}, and Chain-of-Table, all of which follow a modular task-decomposition paradigm: the table QA process is broken down into a series of steps, such as invoking a LLM or embedding-based retriever to identify relevant rows and columns, followed by code generation and program execution to extract data and complete the reasoning process.

Although these frameworks have demonstrated notable improvements in answer accuracy compared to single LLM-based end-to-end approaches, they still face several challenges—most notably, the inference latency introduced by multiple calls to LLM or embedding retrievers, and the risk of losing critical information due to aggressive table pruning strategies. These issues limit the practicality and deployment efficiency of such systems \citep{shao2025ai, an2025ai}.

\section{Methods}\label{sec3}
\subsection{Overview}
\begin{figure*}[htbp]
    \centering 
    \includegraphics[width=0.9\textwidth]{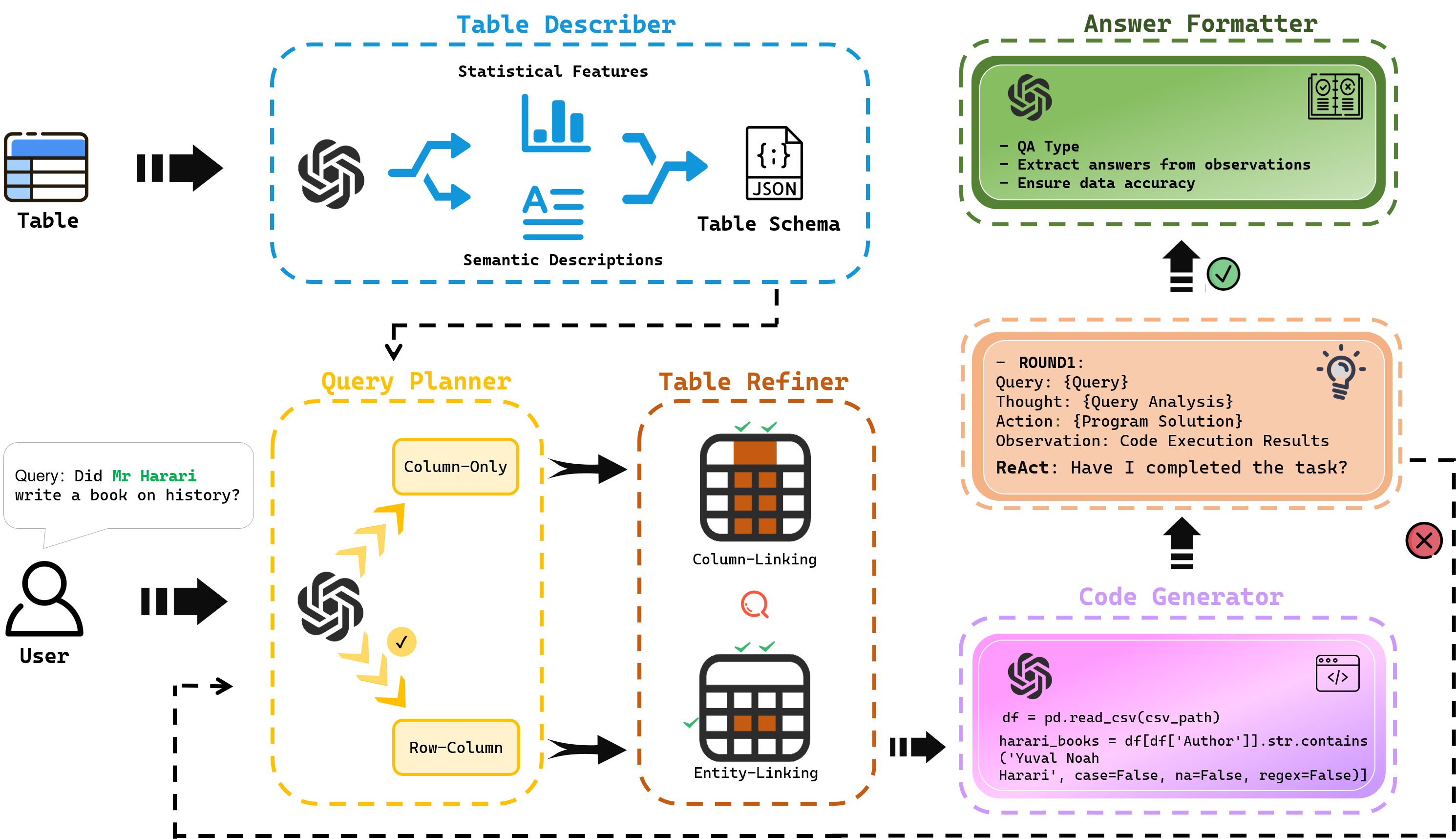}
    \caption{The overview of our TableZoomer framework for table question answering.}
    \label{fig_overall}
\end{figure*}

In this paper, we propose TableZoomer, a specialized agent framework designed for TQA. As illustrated in Fig \ref{fig_overall}, the framework consists of five functional roles powered by LLMs: Table Describer, Query Planner, Table Refiner, Code Generator, and Answer Formatter. These roles collaborate to execute question-answering reasoning. Furthermore, we integrate the ReAct paradigm as the core ``Commander in Chief" of reasoning, empowering the framework with incremental self-reflection and decision-making capabilities. The synergistic operation of these roles constitutes one \textit{Thought-Action-Observation} cycle within ReAct. The overall algorithmic procedures of TableZoomer reasoning workflow are outlined in Algorithm \ref{algo1}.

\subsection{Core components of TableZoomer}
\begin{figure*}[htbp]
    \centering
    \includegraphics[width=\linewidth, scale=5]{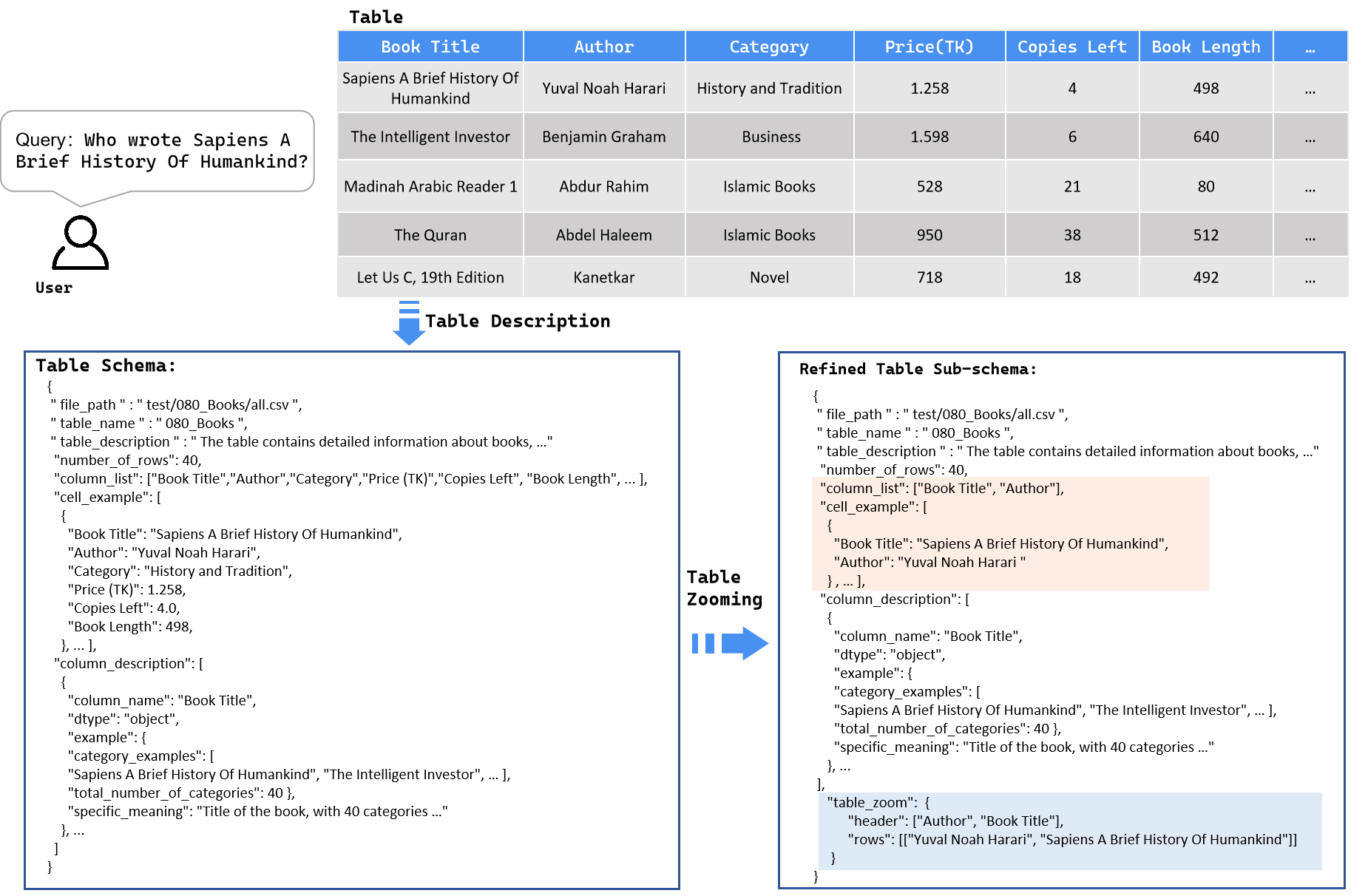}
    \caption{The illustration of table description and table zooming.}
    \label{fig_gen_table_schema}
\end{figure*}

\textbf{Table Describer}.
Prior to the reasoning phase, our framework employs a Table Describer to generate a comprehensive schema for each input table. This schema integrates the table's statistical characteristics, representative examples, and semantic annotations. The process begins by parsing the spreadsheet file using Pandas toolkit. For each column, we extract its data type alongside statistical properties—such as maximum, minimum, mean, and median values for numerical data; unique categories and the most frequent items for categorical data. To provide contextual samples, $K$ randomly selected cell values per column are included as column-wise examples, while $J$ randomly chosen full-row records serve as row-wise examples. Subsequently, these extracted metadata constructs an initial schema to prompt LLM to generate textual descriptions for both the overall table content and individual columns, thereby enriching semantic understanding. The finalized global table schema is organized as structured JSON data. Fig~\ref{fig_gen_table_schema} illustrates the table description.

Notably, schema generation is performed only once per table, with the resulting schema supporting repeated question-answering operations over the same table. The Schema-based representation method substantially enhances TableZoomer's scalability for large-scale tables.

\textbf{Query Planner}.
Table is structured as a two-dimensional data format, where columns typically represent distinct features, and rows correspond to individual data samples. To efficiently handle diverse user queries, we categorize them into two types based on their relationship to the table structure: column-only queries and row-column queries. Column-only queries (e.g., ``calculate the average sales per publisher") are resolved solely using aggregate statistics derived from specific columns. Conversely, row-column queries (e.g., ``calculate the sales for all books by author \textit{Mo Yan}") necessitate the simultaneous identification of relevant rows meeting specific conditions and associated columns.
The core function of the Query Planner is to parse user queries based on the global table schema and direct them to the appropriate processing logic. Specifically, we employ an LLM guided by CoT prompting to decompose the original query into a sequence of specific, atomic sub-queries. Each sub-query is then classified. For column-only sub-queries, the module extracts the associated column names. For row-column sub-queries, it parses both the relevant column names and the corresponding row filtering conditions. Through this process, potential evidence relevant to the user's query is precisely located.

\textbf{Table Refiner}. In table reasoning, directly inputting the entire table often introduces redundant data, which dilutes effective features and impairs key information localization. To address this, building upon key information localization achieved by the Query Planner, we introduce the Table Refiner. Its core objective is to precisely refine the table schema, retaining only information highly relevant to the user query. From the perspective of feature engineering, this process is equivalent to feature pruning. The refinement comprises three key operations:
\begin{itemize}
\item Column Selection: Identifies and retains columns relevant to the query. Leveraging the query expansion information provided by the Query Planner, it filters out irrelevant columns, thereby significantly reducing noise and focusing subsequent reasoning on key district. This process lays the foundation for Table Zooming.
\item Entity Linking: Aligns entities mentioned in the query with corresponding cell values in the table using the row filtering criteria output by the Query Planner. For instance, ``Mr Harari" in the query is linked to ``Yuval  Noah Harari" in the table cell. Specifically, all elements within the relevant columns are programmatically read, and the Longest Common Subsequence (LCS) algorithm is applied to retrieve candidate entries whose overlap rate with the query entity exceeds a threshold of 0.6.

\item Table Zooming: Dynamically focuses the schema based on query type. For column-only queries, it retains relevant columns and optionally incorporates more sample information, while irrelevant columns are removed. For row-column queries, Entity Linking is additionally applied beyond Column Selection, linking the query entity to cell value. This entity clarification is provided as supplementary knowledge to subsequent reasoning steps.
\end{itemize}
As shown in Fig~\ref{fig_gen_table_schema}, the original global table schema is significantly compressed into a highly focused, query-relevant sub-table schema after refinement. This approach enhances the salience of critical local information, enabling LLM to concentrate their attention on noteworthy regions. It is particularly effective for handling large-scale tables prevalent in industrial scenarios, dramatically reducing input token complexity, mitigating reliance on LLM context window size, and consequently improving overall reasoning efficiency.

\textbf{Code Generator}. TableZoomer centers on programmatic execution to derive precise results. It utilizes refined table schema to prompt an LLM to generate PoT solutions. The generated code is subsequently executed within an isolated environment to produce verifiable outputs. Moreover, to enhance robustness, we incorporate an error feedback mechanism: when code execution fails, this mechanism feeds the code along with corresponding error traces back to the LLM, prompting it to re-generate corrected code for self-correction. Compared to purely textual reasoning approaches, this program-assisted solution effectively alleviates numerical hallucinations during multi-step data acquisition and processing, establishing it as an effective strategy for table reasoning.

\textbf{Answer Formatter}.
The Answer Formatter serves as the final module, tasked with producing the formatted response. It incorporates and summarizes intermediate reasoning thoughts and observations throughout the iterative reasoning process. To accommodate strict answer formatting requirements in QA tasks—such as generating complete natural language sentences or purely numerical outputs—users can specify the desired output format via this module.

\subsection{ReAct Paradigm}
Inspired by the ReAct framework, we design an iterative reasoning paradigm tailored to the TQA tasks. As illustrated in Fig \ref{fig_overall}, this paradigm establishes a cyclic reasoning workflow: Within each iteration, the \textit{Thought} corresponds to the query analysis performed by Query Planner; the \textit{Action} involves program ideas generated by Code Generator; and the \textit{Observation} captures the execution outcomes of programs. Following each cycle, the system evaluates whether the current reasoning state can yield the final answer. If so, the result is forwarded to the Answer Formatter; otherwise, a new query is generated and the process proceeds to the next iteration.

\renewcommand{\algorithmicrequire}{\textbf{Input:}}
\renewcommand{\algorithmicensure}{\textbf{Output:}}
\begin{algorithm}
\caption{Reasoning workflow of TableZoomer framework}\label{algo1}
\begin{algorithmic}[1]
\Require {Table $T$, Question or statement $Q$, React round $K$, React thinking records $R$, LLM $M$ and Python Interpreter $P$.}
\Ensure {Answer $A$ to Q.}
\State Initial Table Schema $S \leftarrow $ TableDescriber($T, M$), think round $i = 1$.
\While{$i \le K $}
    \State Query analysis $U \leftarrow $ QueryPlanner($S, Q, M$)
    \State Relevant colomns $C \leftarrow$ getAllRelevantCols($U$)
    \State Refined table schema $S' \leftarrow $ removeRedundantCols($S, C$)
    \For{$q \leftarrow $ getSubQuery($U$) }
        \If {getQueryType($q$) is row-column}
            \State $c$=getRelevantCols($U, q$)
            \State $v'=LCS(e, v)$        $e$: entity extracted from $q$, $v$: values of specific column $c$
            \State Add entity clarification information ($e, v'$) to $S'$
        \EndIf
    \EndFor
    \State $O \leftarrow $ CodeGenerator($Q, S', M$)
    \State $Y \leftarrow $ P.resetEnv().execute($O$)
    \State Append $[Q, U, O, Y]$ to $R$.
    \State $ret \leftarrow $ ReAct($S, Q, R, M$)
    \If{I have completed the task $\in$ ret}
        \State update $Q \leftarrow $ getFurtherQuery($ret$)
        \State $ i=i+1$
    \Else
        \State break
    \EndIf
\EndWhile
\State $A \leftarrow $AnswerFormatter($Q, R, M$)
\State \Return $A$

\end{algorithmic}
\end{algorithm}

\section{Experiment Setup}
\subsection{Datasets}

We evaluate TableZoomer on three popular TQA benchmark datasets: DataBench~\citep{oses-etal-2024-databench}, TableBench~\citep{wu2025tablebench}, and WikiTableQA~\citep{wikitqa-2015}. 
DataBench is unique for its ultra-large tables, whereas the other two cover small to medium sized tables. Table~\ref{tab:dataset_stats} summarizes the statistics of the test sets drawn from these public datasets. Below is the detailed introduction:

\begin{itemize}
\item \textbf{DataBench} is an English benchmark dataset comprising 80 real-world tables. It spans five major domains: business, health, social, sports, and travel, with a 49/16/15 split for training, development, and test sets, respectively. The dataset features five question-answering types: boolean, category, number, list[category], and list[number]. The test set alone encompasses 522 question-answer pairs across 15 datasets within the five domains.\\

\item \textbf{TableBench} is a comprehensive benchmark dataset. The dataset includes 3,681 tables, covering 20 topics such as finance, competitions, sports, and science. It comprises 886 question-answer pairs across 18 subcategories, which are grouped into four main categories: Fact Verification, Numerical Reasoning, Data Analysis, and Visualization. We select the Fact Verification and Numerical Reasoning tasks for evaluating, as they are more aligned with the TQA task. The test set includes 96 entries for Fact Verification and 397 entries for Numerical Reasoning.\\

\item \textbf{WikiTableQA} is the first complex question answering dataset specifically designed for semi-structured tables. It features 22,033 complex question-answer pairs and 2,108 HTML tables sourced from Wikipedia. The questions in this dataset involve various logical operations, such as comparisons, aggregations, and arithmetic operations, requiring models to perform diverse logical and semantic operations. The test set alone comprises 4,344 question-answer pairs and 421 tables.

\end{itemize}

\begin{table}[htbp]
\centering
\caption{Statistics of three benchmark datasets.}
\label{tab:dataset_stats}
\renewcommand{\arraystretch}{1.2}
\begin{tabular}{lccc}
\noalign{\hrule height 1pt}  
\textbf{Dataset} & \textbf{\# Tables} & \textbf{ \# Questions} & \textbf{\# Avg Table Cells} \\
\hline
DataBench (Test) & 15 & 522 & 513359 \\
TableBench (Test) & 493 & 493 & 110 \\
WikiTableQA(Test) & 421 & 4344 & 166 \\
\noalign{\hrule height 1pt}  
\end{tabular}
\end{table}

\subsection{Evaluation}
To ensure the fairness and reproducibility, we adopt the official evaluation implementations provided by each benchmark dataset, and uniformly use Exact Match(EM) Accuracy as the evaluation criterion.

The EM Accuracy metric assesses whether the model's prediction exactly matches the reference answer, with no allowance for partial correctness. Prior to comparison, both prediction and reference are typically normalized by lowercasing, removing punctuation, and trimming whitespace. Only predictions that are exactly the same as the ground-truth answer after normalization are considered correct. This strict criterion is particularly suitable for TQA tasks with verifiable standard answers, where precision is essential.

\subsection{Baselines}
To comprehensively evaluate performance, we select a set of representative open-source LLM as baseline models, covering a variety of scales and architectures. These include Qwen3 series \citep{yang2025qwen3technicalreport}, Qwen2.5-32B~\citep{qwen2025qwen25technicalreport}, LLaMA3.1-70B~\citep{grattafiori2024llama3herdmodels}, QwQ-32B~\citep{qwq32b}, and TeleChat2.5-35B~\citep{he2024telechattechnicalreport,wang2024telechat,li2024teleflmtechnicalreport, wang-etal-2024-telechat,DBLP:journals/corr/abs-2401-03804,wang2025technicalreporttelechat2telechat25}, all of which have demonstrated strong capabilities in table reasoning tasks and reflect the current state-of-the-art  (SOTA). 

For prompt-based methods, we focus on two widely adopted and empirically effective prompting strategies: TCoT and PoT. In parallel, we compare with a range of representative framework-based baselines, encompassing both traditional approaches and advanced systems.

\begin{itemize}
\item \textbf{Binder} adopts a program-guided reasoning paradigm by translating natural language questions into executable code snippets, such as SQL or Python. These programs are executed by external engines to retrieve or verify tabular information with high precision. The method provides symbolic-level reasoning capabilities that complement language models.

\item \textbf{Dater}  decomposes complex questions into multiple subtasks, each addressed individually through table-based or textual reasoning. The model then sequentially integrates these local reasoning results using a hierarchical chain structure. This stepwise composition facilitates robust reasoning over multi-faceted queries.

\item \textbf{Chain-of-Table} constructs a multi-step reasoning process by iteratively recording intermediate results in table form. At each step, the model generates explicit operations such as filtering, aggregation, or sorting, and updates the table accordingly. It enables a dynamic evolution of reasoning steps grounded in structured tabular representations.

\item \textbf{TableRAG} combines schema and cell retrieval with query expansion to efficiently extract relevant table content. By avoiding full-table encoding and selecting only salient column names and cell values, it enables scalable and token-efficient table reasoning with LLMs.

\item \textbf{PoTable}~\citep{mao2025potablesystematicthinkingstageoriented} builds hierarchical indices based on the internal structure of the table and operates within a retrieval-reasoning loop. It leverages pointer networks or similarity-based matching to locate relevant cells and performs iterative inference within localized subtable regions. This approach enhances controllability and scalability in reasoning over large tables.
\item \textbf{TableMaster} enhances table understanding by semantically verbalizing tabular content and applying an adaptive strategy that dynamically switches between neural and symbolic reasoning. Depending on the nature of the query, it selects the most suitable reasoning pathway. This flexibility allows it to handle diverse and semantically complex table structures effectively.
\item \textbf{TabSQLify}~\citep{nahid-rafiei-2024-tabsqlify} converts natural language questions into SQL queries to retrieve a focused subtable from the original table. The subtable is then passed to a LLM for final inference. This two-stage process reduces context length and improves reasoning efficiency, especially for large or information-dense tables.
\end{itemize}

\subsection{Implementation Details}
For Qwen3 series LLMs, we adopt the think mode by default, unless explicitly specified as ``nothink". All LLMs are evaluated with a decoding temperature of 0 during inference to ensure deterministic outputs. In both the TCoT and PoT prompting approaches, the maximum sequence length is set to 32,768. For ReAct reasoning, the number of reasoning rounds is capped at 5.

Under both PoT and TCoT paradigms, tables are textualized in Markdown format, a widely used verbalization method for tabular data. To reduce model bias and handle fine-grained reasoning requirements, we design carefully crafted prompts that incorporate few-shot CoT demonstrations and structured output constraints. For the sake of fairness, we format program execution results of PoT through the Answer Formatter to produce final answers. 

We include several prompts we used in the Appendix~\ref{appendix_prompt}.

\section{Results and Analysis}
\subsection{Main Results}

\subsubsection{Results on DataBench and TableBench datasets}

\begin{table}[h]
\caption{Experimental results on DataBench dataset. The LLMs used in the experiment are either chat or instruction model versions. The Qwen3 series has enabled thinking mode.}
\centering
\small
\setlength{\tabcolsep}{2pt}
\begin{tabular*}{1\textwidth}{ll|ccccc}
\toprule
\textbf{Method} & \textbf{Avg [\%]} & boolean & category & number & list[category] & list[number] \\
\midrule 
Qwen3-8B & 67.82 & 76.74 & 62.16 & 74.36 & 55.56 & 58.24 \\
\hspace{1em} w/ TableZoomer & 87.16\scriptsize{($\uparrow$ 19.34)} & 92.25 & 91.89 & 85.26 & 83.33 & 82.42 \\
\midrule
Qwen3-14B & 76.25 & 86.82 & 71.62 & 74.36 & 65.28 & 75.92 \\
\hspace{1em} w/ TableZoomer & 87.93\scriptsize{($\uparrow$ 11.68)} & 94.57 & 85.14 & 84.62 & 83.33 & 90.11 \\
\midrule
Qwen2.5-32B & 75.48 & 87.60 & 74.32 & 69.87 & 73.61 & 70.33 \\
\hspace{1em} w/ TableZoomer & 88.12\scriptsize{($\uparrow$ 12.64)} & 95.35 & 87.84 & 84.62 & 83.33 & 87.91 \\
\midrule
Qwen3-32B & 80.08 & 93.02 & 68.92 & 80.13 & 79.17 & 71.43 \\
\hspace{1em} w/ TableZoomer & 90.42\scriptsize{($\uparrow$ 10.34)} & 97.67 & 94.59 & 87.18 & 83.33 & 87.91 \\	
\midrule
QwQ-32B & 77.97 & 89.15 & 78.38 & 78.21 & 62.5 & 73.63 \\
\hspace{1em} w/ TableZoomer & 88.12\scriptsize{($\uparrow$ 10.15)} & 96.9 & 86.49 & 85.9 & 75 & 91.21 \\
\midrule
TeleChat2.5-35B & 70.31 & 82.17 & 72.97 & 66.03 & 59.72 & 67.03 \\
\hspace{1em} w/ TableZoomer & 83.72\scriptsize{($\uparrow$ 13.41)} & 96.12 & 75.68 & 83.33 & 77.78 & 78.02 \\

\midrule
Llama3.1-70B & 76.44 & 86.82 & 71.62 & 69.87 & 72.22 & 80.22 \\
\hspace{1em} w/ TableZoomer & 84.29\scriptsize{($\uparrow$ 7.85)} & 92.25 & 87.84 & 82.05 & 73.61 & 82.42 \\
\bottomrule
\end{tabular*}
\label{table:exp_main_res}
\end{table}

\begin{table}[h]
\belowrulesep=0pt
\aboverulesep=0pt
\renewcommand{\arraystretch}{1.3}
\caption{Experimental results on Fact Checking and Numerical Reasoning tasks of TableBench dataset. The LLMs used in the experiment are instruction model versions. The Qwen3 series has enabled think mode.}\label{tab-tablebench}
\begin{tabular*}{0.8\textwidth}{@{\extracolsep\fill}cccc}
\bottomrule
\multirow{2}{*}{Method} & \multirow{2}{*}{Base Model} & \multicolumn{2}{c}{TableBench [\%]}  \\ 
\cmidrule{3-4}
 &  & Fact Checking & Numerical Reasoning \\
\midrule
\multirow{3}{*}{PoT-only} & Qwen3-8B & 54.17 & 54.41 \\
 & Qwen3-32B & 61.46 & 54.66 \\
 & Llama3.1-70B & 59.05 & 34.04 \\
\midrule
\multirow{3}{*}{TableZoomer} & Qwen3-8B & 79.17\scriptsize{($\uparrow$ 25)} & 66.25\scriptsize{($\uparrow$ 11.84)} \\
 & Qwen3-32B & 83.33\scriptsize{($\uparrow$ 21.87)} & 71.28\scriptsize{($\uparrow$ 16.62)} \\
 & Llama3.1-70B & 81.25\scriptsize{($\uparrow$ 22.2)} & 68.26\scriptsize{($\uparrow$ 34.22)} \\
\toprule
\end{tabular*}
\end{table}

We evaluate TableZoomer against baseline LLMs using PoT prompting on the DataBench and TableBench datasets. As shown in Table \ref{table:exp_main_res} and Table \ref{tab-tablebench}, TableZoomer consistently enhances performance across all LLMs significantly, with particularly pronounced improvements observed on smaller-scale models, demonstrating our framework’s superiority.
Among all evaluated LLMs, TableZoomer integrated with Qwen3-32B achieves the highest accuracy. Simultaneously, the Qwen3-8B driven TableZoomer delivers notable competitive results. It elevates absolute accuracy by 19.34\%, 25\%, and 11.84\% points on the DataBench and the Fact Checking and Numerical Reasoning tasks of TableBench, respectively. These findings validate the scalability of our framework, which maintains significant performance improvements across varying table sizes.

Besides, from the experimental results of Qwen3 model series on DataBench, we observe that TableZoomer effectively narrows the performance gap between models of different scales while substantially improving the cost-effectiveness of smaller models. Notably, the TableZoomer-enhanced Qwen3-8B even outperforms the baseline Qwen3-32B model, highlighting TableZoomer’s potential for enabling smaller models to achieve high performance under resource-constrained scenarios. It's a crucial advantage for real-world deployment.

\subsubsection{Results on WikiTableQA dataset}
We compare TableZoomer using LLama3.1-70B as the LLM solver against advanced programming-assisted fremawork on small-scaled WikiTableQA dataset. Results in Table \ref{tab-wikiqa-tabfact} show that TableZoomer achieves competitive performance on WikiTableQA, slightly trailing the top baseline TableMaster by 1.43\% while outperforming the second ranked PoTable by 10.96\%. 
The results confirm TableZoomer’s ability to adapt effectively to tables of diverse sources and scales, exhibiting strong generalization and scalability.

\begin{table}[h]
\belowrulesep=2pt
\aboverulesep=2pt
\renewcommand{\arraystretch}{1.2} 
\renewcommand{\thefootnote}{\alph{footnote}} 
\caption{Accuracy and inference cost comparison on WikiTableQA dataset.}
\label{tab-wikiqa-tabfact}
\begin{tabular}{@{}lcc@{}}
\toprule
Method  & WikiTableQA & Inference times per sample \\
\hline
Binder & 50.51 & \scriptsize{Number of self-consistent executions.}  \\
Dater & 43.53 &  4 \\
TabSQLify &  55.78 & 2\\
Chain-of-Table &  62.22 & \scriptsize{Number of planning iterations.} \\
TableRAG \footnotemark[1] & 57.03 & 1+\scriptsize{Number of sub-queries.}  \\
PoTable & 65.56 & $\geq 8$ \\
TableMaster  & \textbf{77.95} & $\geq 6$ \\
TableZoomer (Ours)  & \underline{76.52} & $\geq 5$ \\
\botrule
\end{tabular}

\footnotetext{In addition to the results of TableRAG and our proposed TableZoomer framework, results for other methods are sourced from \cite{cao2025tablemasterrecipeadvancetable}. TableRAG employs GPT-3.5-turbo as its LLM solver, while other methods there utilize LLaMA3.1-70B-instruct as their LLM solver.}
\footnotetext[\textsuperscript{a}]{{The result is source from \cite{chen2024tablerag}.}}

\end{table}

\subsubsection{Comparison with TCoT and PoT reasoning methods}

\begin{figure}[h]
\centering
\includegraphics[width=0.8\textwidth]{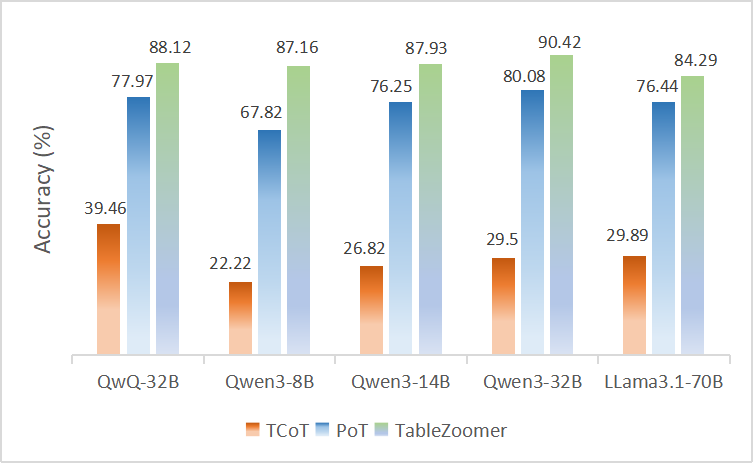}
\caption{Accuracy comparison with TCoT and PoT prompting baselines on DataBench dataset.}\label{fig_comparison_tcot_pot_tablezoomer}
\end{figure}

To investigate the efficacy of different table reasoning paradigms, we compare the performance of TCoT, PoT, and TableZoomer on the Datebench dataset. As shown in Fig  \ref{fig_comparison_tcot_pot_tablezoomer}, PoT method yields a substantial performance leap over TCoT, achieving absolute accuracy gains exceeding 40\% across various baseline models. This further underscores the effectiveness of programming-assisted strategies for reasoning over large-scale tables. Building upon PoT, TableZoomer integrates LLM-based Collaborative reasoning with table schema representation and zooming operations to further deliver significant performance improvements.

\subsubsection{Performance Analysis Across Table Sizes}

\begin{figure}[h]
    \centering
    \includegraphics[width=0.6\textwidth]{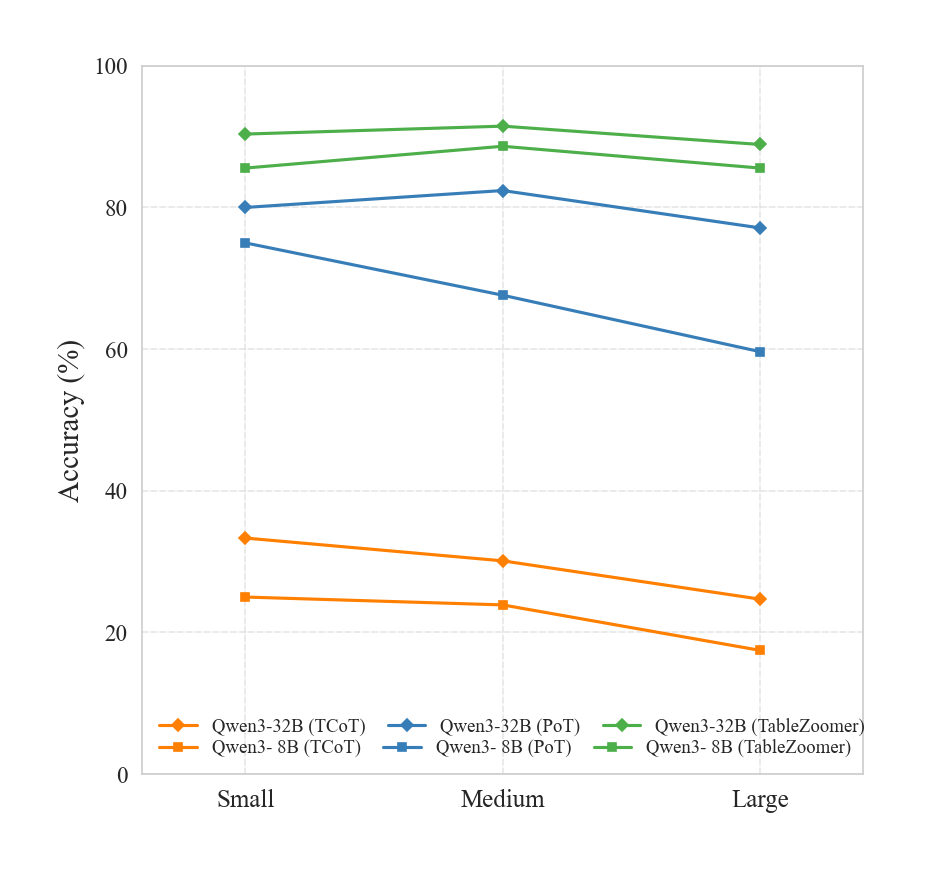}
    \caption{Accuracy comparison with TCoT and PoT prompting baselines across table sizes on DataBench dataset.}
    \label{table_size_fig}
\end{figure}

We sort tables in the DataBench dataset in descending order based on table cell count and categorize them into large (avg. 1,519,065 cells), medium (avg. 18,131 cells), and small (avg. 2,882 cells) subsets. Fig \ref{table_size_fig} reveals that TCoT suffers severe performance degradation as table size increases. The code-based PoT approach consistently outperforms TCoT across table sizes, validating its efficacy for tabular data processing and showing its potential for large-scale tables. Remarkably, TableZoomer demonstrates exceptional scalability and robustness, exhibiting minimal performance degradation with increasing table size. This is primarily attributed to the core TableDescriber and TableRefiner components, which excels at locating and focusing on necessary information from large tables.

\subsubsection{Effects of different table representations format}

\begin{figure}[h]
\centering
\captionsetup{justification=centering}
\includegraphics[width=1\textwidth]{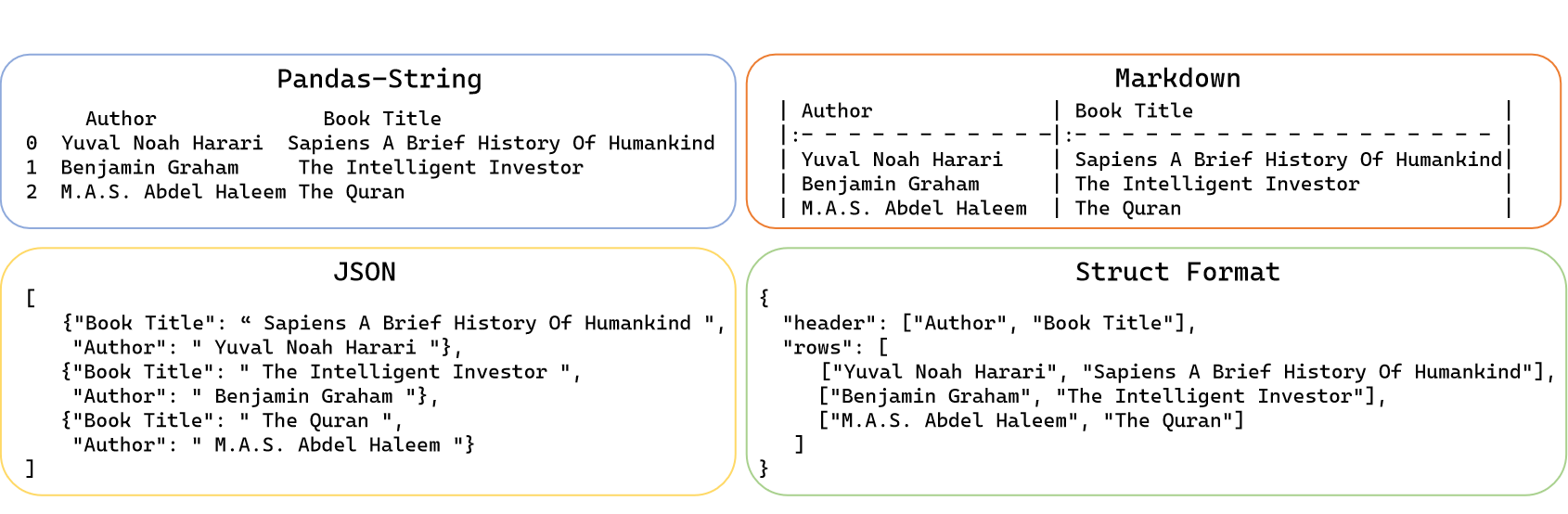}
\caption{Four different table representation formats.}
\label{fig_diff_repr}
\end{figure}

To enable LLMs to accurately understand tabular data, it is essential to transform the data from a two-dimensional structured format into a one-dimensional textual sequence \citep{Deng2024TextTupleTableTI}. During this transformation, the row-column relationships and hierarchical structure of the table must be effectively preserved; as such, different serialization strategies have a critical impact on the model’s reasoning performance \citep{Zhang2024ASO}.  In light of this, we evaluate four widely adopted table representation formats to compare the performance of PoT and our proposed TableZoomer on the DataBench dataset, including Pandas-String, Markdown, JSON, and Struct-Format, as show in Fig~\ref{fig_diff_repr}.

The comparison results are listed in Table~\ref{tab:repr_format_comparison}. For PoT approach, Qwen3-8B achieves its best performance when tables are serialized in JSON format, followed by Markdown. However, its accuracy exhibits a substantial fluctuation of up to 9.18\% points across the four formats, revealing a high degree of sensitivity to structural encoding. This instability may stem from distributional biases in the model’s training corpus, which affects the ability to generalize across diverse table representations. In contrast, TableZoomer demonstrates remarkable robustness: by leveraging a unified structure-aware module and a dynamic semantic alignment mechanism, it maintains consistently high performance across all formats, with accuracy ranging from 86.59\% to 87.16\% and a maximum variation of only 0.57\% points. These results confirm TableZoomer’s superior generalization capability and resilience to representational variation in tabular inputs.

\begin{table}[h]
\centering
\caption{Comparison of Qwen3-8B based TableZoomer and PoT  across different table formats.}
\begin{tabularx}{0.75\textwidth}{>{\centering\arraybackslash}X
                                  >{\centering\arraybackslash}X
                                  >{\centering\arraybackslash}X}
\toprule
\textbf{Table Format} & \textbf{PoT (\%)} & \textbf{TableZoomer (\%)} \\
\midrule
Pandas-String & 64.37 & \textbf{86.59} \\
Markdown      & 67.82 & \textbf{86.78} \\
JSON          & 73.18 & \textbf{86.97} \\
Struct Format & 63.98 & \textbf{87.16} \\
\bottomrule
\end{tabularx}
\label{tab:repr_format_comparison}
\end{table}

\begin{table}[h]
\belowrulesep=0pt
\aboverulesep=0pt
\renewcommand{\arraystretch}{1.3}
\caption{Ablation results on DataBench dataset with Qwen3-8B based TableZoomer. }\label{tab_ablation}
\begin{tabular*}{0.9\textwidth}{@{\extracolsep\fill}ccccc|c}
\bottomrule
\multirow{2}{*}{Exp ID} & \multirow{2}{*}{Table Schema Rep}\footnotemark[1] & \multicolumn{2}{c}{Table Zooming} & \multirow{2}{*}{ReAct}\footnotemark[2] & \multirow{2}{*}{Acc}  \\ 
\cmidrule{3-4}
  & & Column Selection & Entity Linking &  \\
\midrule

\textit{Ala 1} & &  &  &  & \normalsize{67.82} \\
\textit{Ala 2} & \checkmark & & & & \normalsize{74.33} \\
\textit{Ala 3} & \checkmark & \checkmark & & & \normalsize{84.67} \\
\textit{Ala 4} & \checkmark & \checkmark & \checkmark & & \normalsize{86.40} \\
\textit{Ala 5} &\checkmark & \checkmark & \checkmark & \checkmark & \normalsize{87.16} \\

\toprule
\end{tabular*}
\end{table}

\subsection{Ablation Study}
To further validate the contribution of each component within TableZoomer, we conduct ablation studies on the DataBench benchmark. Experiments are performed using the Qwen3-8B model in thinking mode, with results detailed in Table \ref{tab_ablation}.
The \textit{Ala 1} row reports accuracy of the baseline PoT method. \textit{Ala 2} demonstrates that merely replacing the full-table Markdown text with the Table Schema representation text within the PoT prompt yields an accuracy improvement of 6.51\%, strongly validating the efficacy of the Table Schema representation.
Subsequently introducing the Table Zooming mechanism contribute to a significant leap in accuracy, from 74.33\% to 86.40\%. This unequivocally demonstrates Table Zooming's core role in capturing the hierarchical structure and key information within tables. Removing the Entity Linking module leads to a degradation in accuracy of 1.73\%. Finally, integrating the ReAct reflection-iteration mechanism yield a modest yet consistent accuracy gain on DataBench.
These experiments clearly illustrate that the synergistic effect of all components collectively drives the continuous optimization of model performance. Among them, the Table Schema representation and Table Zooming mechanism are identified as particularly critical roles.

\subsection{Efficiency Analysis}
\begin{figure}[h]
\centering
\includegraphics[width=0.7\textwidth]{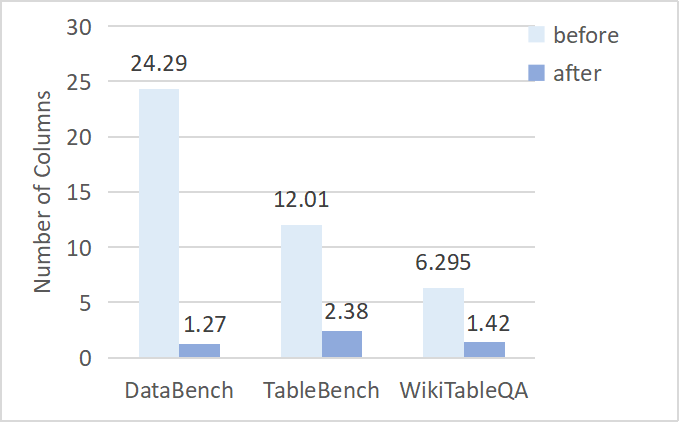}
\caption{Changes in the number of column features contained in the table schema before and after the table zooming operation.}\label{fig_changes_of_column_nums}
\end{figure}

\subsubsection{Input token reduction}
Let $M$ and $N$ denote the number of rows and columns in a table, respectively. The token complexity of a flattened table representation obtained by verbalizing all cells is $O(M \times N)$. In contrast, the token complexity of table schema representation employed by TableZoomer, which describes the column structure (including semantics and exemplar values), is $O(N)$, representing a reduction of orders of magnitude compared to $O(M \times N)$.

Table Refiner performs query-aware table zooming, dynamically selecting a relevant subset of columns based on the input query. This reduces the number of columns from N to $N'$ (where $N' < N$), further optimizing the input token complexity. As shown in Fig \ref{fig_changes_of_column_nums}, on the DataBench, TableBench, and WikiTableQA datasets, the column selection operation achieves compression ratios of 5.2\%, 19.8\%, and 22.6\% of the original column counts. Crucially, TableZoomer preserves high accuracy performance, validating that the refined sub-schema effectively retains the critical information necessary for question answering.

TableZoomer achieves promising efficiency in processing large-scale tables through dual-stage optimization mechanism: (1) Table Schema Representation mitigates token inflation along the row dimension, and (2) Query-Aware Column Selection within the Table Refiner addresses the challenge posed by the column dimension. Consequently, TableZoomer demonstrates significant efficiency advantages for tables with either numerous rows or columns.

\subsubsection{Reasoning efficiency}
The computational resource and time consumption associated with LLM dominate the reasoning efficiency of LLM-based TQA framework \citep{Min2024ExploringTI}. We evaluate reasoning efficiency of different frameworks by measuring the times of LLM inference required.
Binder employs a self-consistency sampling strategy to enhance performance, requiring multiple LLM inferences to generate and aggregate multiple Neural-SQL queries. In contrast, frameworks such as Chain-of-Table, PoTable, and TableMaster rely on iterative reasoning strategies where inference counts typically scale with problem complexity. Specifically, Chain-of-Table adopts a greedy search strategy during table reasoning process, and the inference count equals the total number of planning and execution operations. PoTable involves file distinct stages, and its inference count depends on the number of steps within the planning chain for each of the latter four stage, typically requiring no fewer than eight LLM invocations. TableMaster encompasses multiple steps including potentially repeated column/row lookups, generally demanding at least six LLM calls. TablaRAG, utilizing hierarchical prompt reasoning, sees its invocation count determined by the number of sub-queries generated during tabular query expansion, typically ranging from three to five. Dater executes four fixed yet intricate steps, resulting in exactly four LLM inferences.

TableZoomer's core reasoning workflow requires a minimum of five LLM invocations spanning the Table Describer, Query Runner, Code Generator, Answer Formatter, and ReAct reflection. The innovative design significantly enhances efficiency: Table Describer executes only once per spreadsheet, with outputs being reusable across subsequent multi-turn QA or reflection rounds to eliminate redundant computation. Meanwhile, the reflection mechanism activates only when necessary, avoiding rigid multi-round iterations. As demonstrated in Table \ref{tab-wikiqa-tabfact}, TableZoomer achieves a remarkable balance between computational efficiency and answer quality.

\subsection{Case Analysis}
For programming-assisted TQA methods, errors primarily stem from question/table comprehension deviations, code generation errors (e.g., column name mismatches, incorrect value conditions, syntax errors), and answer formatting issues. To gain insight into practical effects of our framework, we conduct detailed case studies on the DataBench test set.

\begin{figure}[H]
\centering
\includegraphics[width=0.8\textwidth]{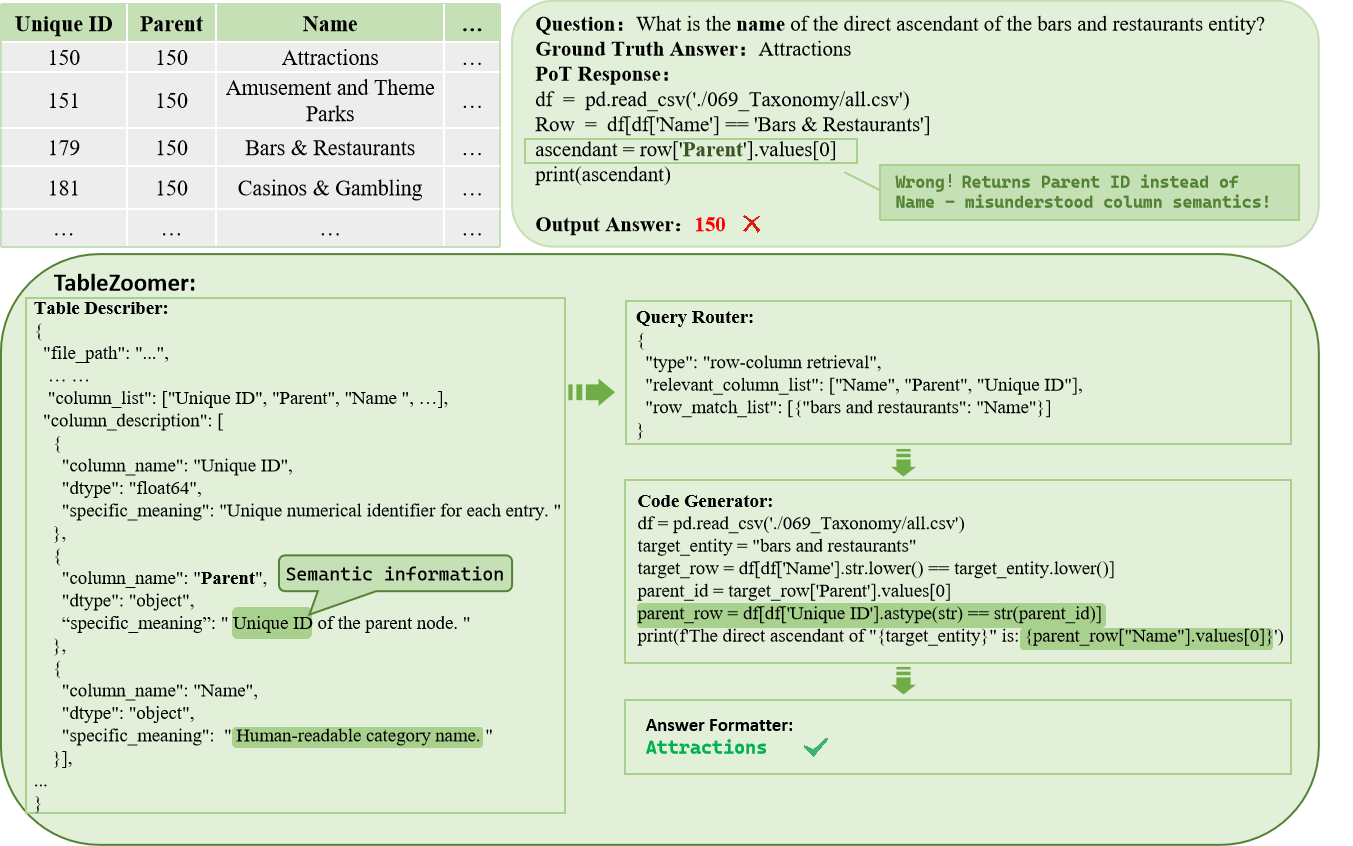}
\caption{Case study of TableZoomer comparing to PoT: column semantic misunderstanding in hierarchical column structure.}\label{misunderstand_query}
\end{figure}

Fig ~\ref{misunderstand_query} demonstrates how PoT can suffer from a misunderstanding of column semantics. In this example, while the model correctly identifies the row corresponding to \textit{Bars} \& \textit{Restaurants}, it returns the \textit{Parent ID} (150) rather than the associated category name (``\textit{Attractions}"). This is due to the model’s failure to recognize that the ``\textit{Parent}'' column functions as a foreign key referencing the ``\textit{Unique ID}''. TableZoomer overcomes this issue by incorporating schema-level semantic annotations via the Table Describer, which clearly specifies the cross-column relationship. The Code Generator is thus able to perform an additional lookup operation, successfully retrieving the human-readable parent category name and yielding the correct answer ``\textit{Attractions}''. 

\begin{figure}[H]
\centering
\includegraphics[width=0.8\textwidth]{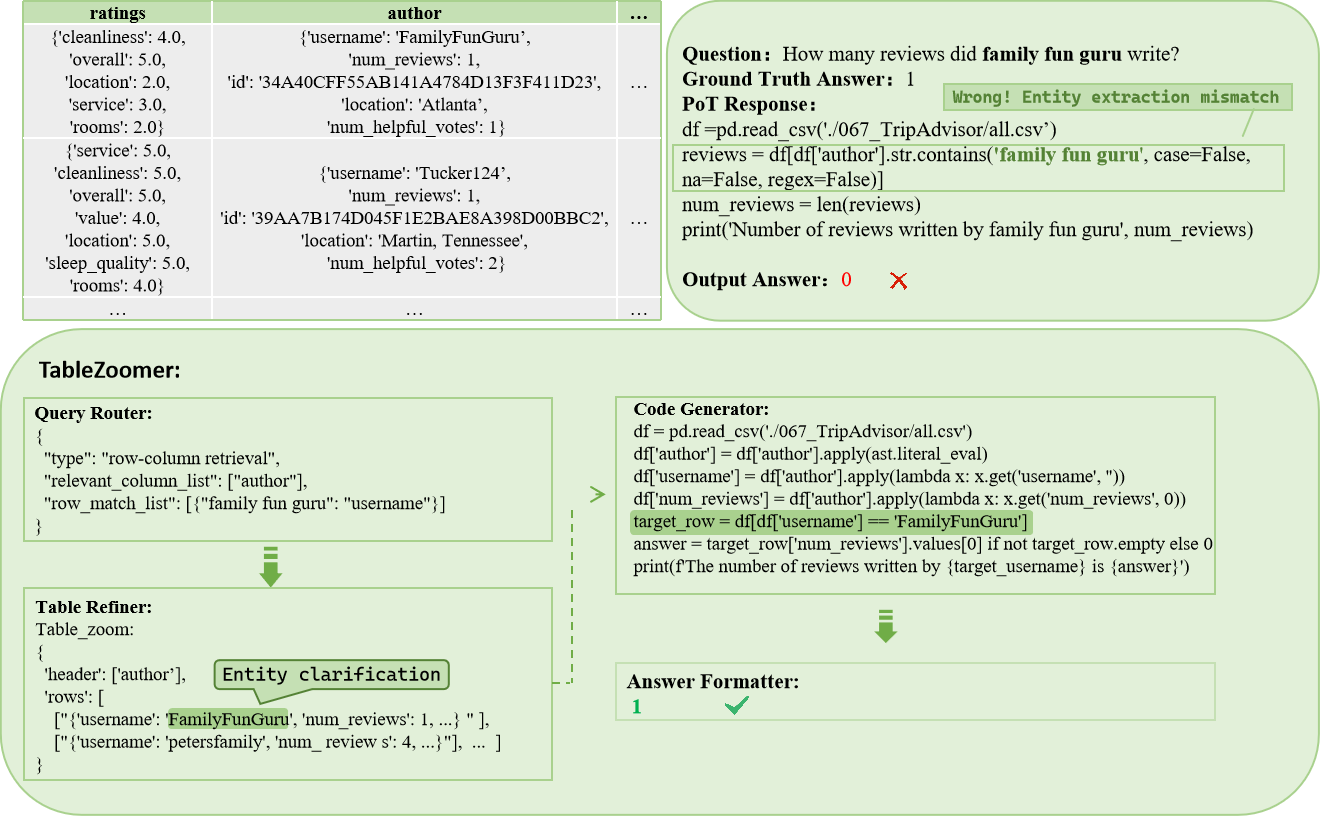}
\caption{Case study of TableZoomer comparing to PoT: entity extraction mismatch due to surface-form variation.}\label{misunderstand_entity}
\end{figure}

Fig~\ref{misunderstand_entity} highlights a key limitation of  PoT in handling entity extraction. In real-world scenarios, the entity mentioned in a question often deviates from its exact surface form in the table, leading to mismatches and subsequent code generation failures. In contrast, TableZoomer first invokes the Query Planner to accurately parse the user intent and locate the relevant columns and rows. Subsequently, the Table Refiner narrows down the global table context to query-specific local views, which are then incorporated into the refined table schema. This enables the Code Generator to retrieve the exact target entity ``\textit{FamilyFunGuru}'' and generate the correct answer through accurate reasoning.

\begin{figure}[H]
\centering
\includegraphics[width=0.8\textwidth]{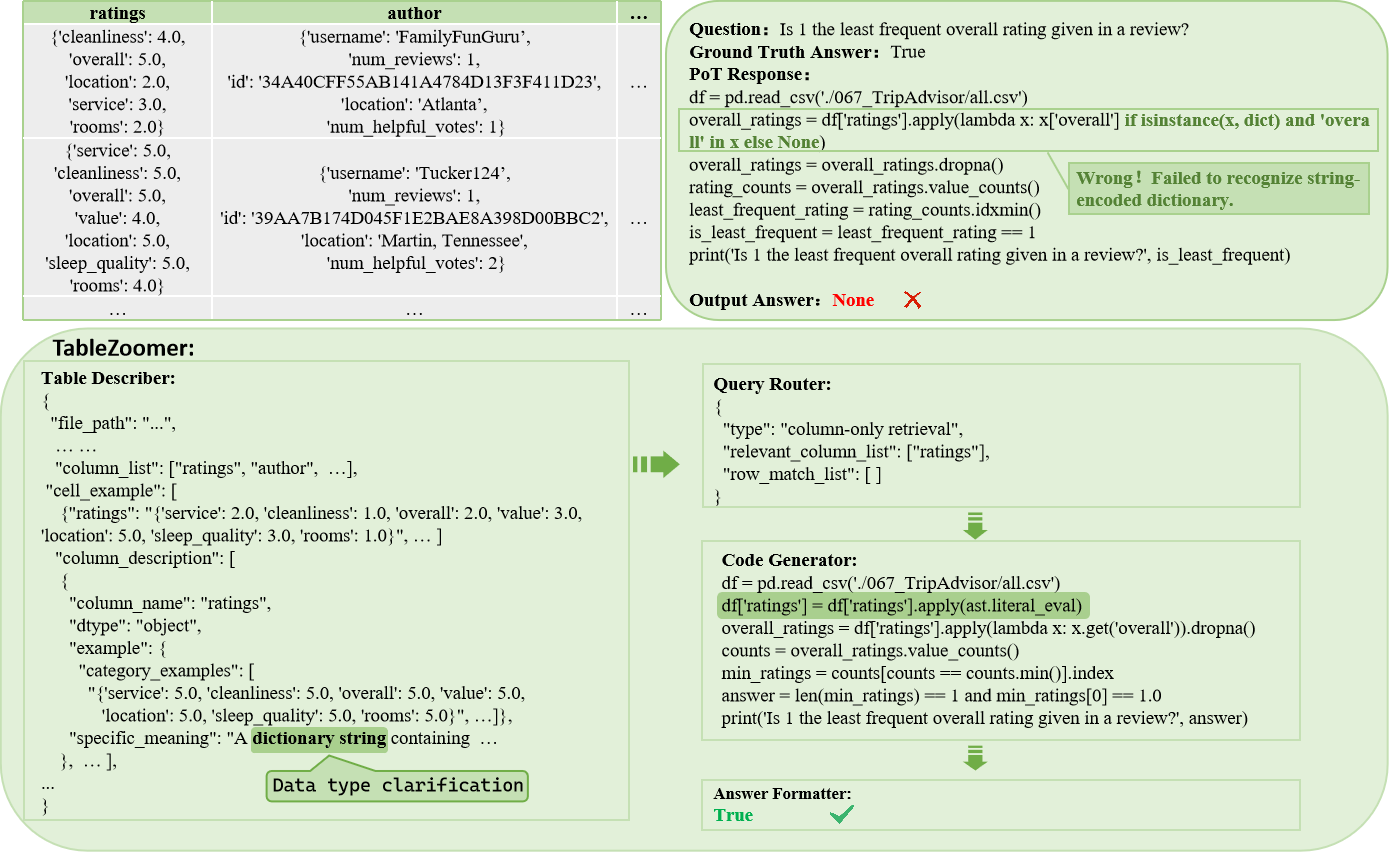}
\caption{Case study of TableZoomer comparing to PoT: failure in parsing string-encoded dictionary fields.}\label{misunderstand_datatype}
\end{figure}

Fig~\ref{misunderstand_datatype} reveals another failure example of PoT, originating from significant data noise in tabular inputs. Specifically, while querying the ``\textit{overall rating}'' from the \textit{ratings} column, PoT fails to parse the field due to its inability to recognize string-encoded dictionaries amidst the noisy data, resulting in access failure and a None output. In contrast, TableZoomer employs the Table Refiner to mitigate noise contamination prior to processing. Through denoising phase, the system successfully recovers interpretable data structures. Subsequently, the Code Generator applies \texttt{ast.literal\_eval} to parse structured dictionaries, enabling correct extraction of ``\textit{overall}'' ratings and ultimately producing the valid answer ``\textit{True}''.

\section{Conclusion}\label{sec13}

This paper introduces TableZoomer, an LLM-powered, programming-based agent framework for TQA tasks. To address the characteristics of tabular data, we adopt the table schema as representation, substantially reducing token complexity and enhancing reasoning efficiency for large-scale tables. The framework employs a query-aware table zooming mechanism to refine the schema into a focused, high-information-density sub-schema, effectively eliminating redundancy while further reducing computational costs. Furthermore, we implement the ReAct reasoning paradigm with iterative workflows to facilitate incremental thinking and reflection. Extensive experimental results validate TableZoomer's superiority, demonstrating high performance and scalability across various table sizes, with especially significant advantages for large-scale tables.

\section{Future Work}
Although TableZoomer demonstrates performance gains through LLM collaboration, the predefined, fixed workflow still constitutes a performance bottleneck \citep{xiong-etal-2025-teleai,Zhao2024TaPERAEF}. In future work, we will explore an agentic framework for table reasoning, employing a central ``brain" module to intelligently select the optimal action path from numerous built-in sub-modules, autonomously constructing flexible workflows that simultaneously achieve the optimal balance between accuracy and reasoning efficiency \citep{ZHAO2025113905,xing-etal-2025-llmsr}. Furthermore, we plan to abstract a universal reasoning scheduler. This scheduler will seamlessly adapt to diverse and complex heterogeneous tabular data while adaptively adjusting the reasoning depth.

In addition, the current TableZoomer framework relies purely on language-based reasoning. With the rapid advances in multi-modal large language models, a promising direction is to extend TableZoomer toward vision-language integration \citep{JIANG20181132,7014257}. By incorporating visual encoders to directly perceive table structures, layouts, and visual elements, the framework could achieve stronger robustness and generalization when handling complex formatting or chart-like tables \citep{1640956,7051244,4633660}. This evolution would move TQA closer to multi-modal understanding, enabling reasoning processes that more closely resemble human cognition.

\begin{appendices}

\section{Prompts}\label{appendix_prompt}

\begin{tcolorbox}[colback=white, colframe=black!30, title=\textbf{TCoT Prompt Example for DataBench}, boxrule=0.6pt, arc=3pt, left=5pt, right=5pt, top=5pt, bottom=5pt,fontupper=\small]

You are an assistant tasked to answer a question about a given table in Markdown format. Use only the information from the table. Respond in a single JSON object with the following fields:
\begin{itemize}[leftmargin=1.2em, itemsep=0pt]
  \item answer: the direct answer derived from the table.
  \item explanation: brief reason for the answer.
\end{itemize}

\textbf{Answer Types:}
\begin{itemize}[leftmargin=1.2em, itemsep=0pt]
  \item Boolean: ``True" or ``False''
  \item Category: a single categorical value
  \item Number: numeric value, e.g., average, max
  \item List: a list of numbers or categories
\end{itemize}

\textbf{Instructions:}
\begin{enumerate}[leftmargin=1.2em, itemsep=0pt]
  \item Do not add any prefix or extra commentary.
  \item Do not infer beyond the given table.
  \item Preserve original formats and values.
\end{enumerate}

\vspace{0.5em}
\textbf{Example}

Question: What is the name of the richest passenger?

\begin{lstlisting}[basicstyle=\normalfont\small,columns=fullflexible]
Table:
|passenger|wealth($)|
|-----|-----|
|value1    |value2    |
\end{lstlisting}

Response:\\
\{ \\
\hspace*{2em}``answer": ``value", \\
\hspace*{2em}``explanation": ``xxx" \\
\}

\textbf{Now let’s start!}

Question: \\
\{question\} \\

Table:

\{csv\_data\} \\

Response:
\end{tcolorbox}

\vspace{1.5em}

\begin{tcolorbox}[
  colback=white, colframe=black!30,
  title=\textbf{PoT Prompt}, boxrule=0.6pt, arc=3pt,
  left=5pt, right=5pt, top=5pt, bottom=5pt,
  breakable,
  fontupper=\small 
]

You are a professional programming assistant designed to utilize the Python package \textbf{pandas} to analyze the table and Response efficient and robust Python code for answering user's Question. The code will read the file from the given `Table\_path` and perform data extraction. \\

You should act in accordance with the following requirements:\\
1. Generate chain-of-thought execution ideas based on the understanding of the table content and the user's Question. Describe in detail the algorithm steps as much as possible, including Question analysis, table data format parsing method and code logic description. \\\

2. Then write Python codes according to your approach to solve the question. The codes need to be concise and easy to understand, and if necessary, add comments for clarification. \\

3. Note that your analysis must be based entirely on the Table data, with special attention to the content and format of the table cells.\\

You should deliberately go through the user's Question, Table\_path and Table and strictly follow the guidelines to appropriately answer the user's Question. You can only output a standardized JSON object, including "code\_thought" and "code", and you are prohibited from outputting any other unnecessary thought processes. Ensure that your Response can be read by json.loads().\\

\textbf{Guidelines}:\\
Thought generation: With the goal of addressing the user's Question, refer to table\_data to generate step-by-step code writing ideas.\\

File Reading: Depending on the table file format and size, efficiently read data from the given table file path (supporting formats such as CSV, Excel) and load it into a Pandas DataFrame. For larger datasets, choose an appropriate method to ensure performance.\\

Special reminders:\\
- The generated code should be robust, including error handling and file format compatibility. It should strictly match the column names mentioned in the user's Question, avoiding irrelevant or mismatched columns.\\

- Unless otherwise specified, please ignore null or empty values.\\

- Pay attention to the wording of the question to determine if uniqueness is required or if repeated values are allowed. Unless otherwise specified, the unique operation (.unique()) is not necessary when sorting or finding the maximum/top/highest/most/lowest/smallest/last (etc) N values in most cases.\\

- Pay attention to the format of example values before you manipulate the data in a certain column. Deeply think about how to correctly parse and extract ill-formed data, Not JUST anomaly capture. \\

- For Boolean problems, it is not necessary to output all elements, only obtain True or False answers, or obtain the first few elements to avoid too much unnecessary output.\\

- The results of mathematical operations must be specific number values, and Scientific notation cannot be used.\\

\textbf{Response Format}:\\
User's Question: Which label has the highest number of products?\\
Response: \\
\{

\hspace{15pt}``code\_thought":``To find the single label with the highest number of associated products, we'll: 1. Parse the \textit{labels\_en} column to extract individual labels; 2. Handle empty lists and string formatting issues; 3. Count occurrences of each label; 4. Identify the label with the highest count.",

\hspace{15pt}``code":``

import pandas as pd\\
def parse\_labels(s):

\hspace{15pt}if s == '[]':

\hspace{30pt}return []

\hspace{15pt}return [label.strip() for label in s.strip('[]').split(',')]

df = pd.read\_csv('all.csv')\\ 
\# Explode the labels into individual rows\\
labels = df['labels\_en'].apply(parse\_labels).explode()\\ 
\# Count occurrences of each label\\
label\_counts = labels.value\_counts()\\
\# Find the label with the highest number\\
most\_common\_label = label\_counts.idxmax()\\
print('the label with the highest number of products', most\_common\_label)

"

\}\\

\textbf{Let's begin!}\\
Now please deliberately go through the following user's Question, Table\_path and Table word by word and strictly follow the above guidelines to appropriately answer the question. You can only output a standardized JSON object, including ``code\_thought" and ``code", and you are prohibited from responsing without any prefix words or explanations. Ensure that your Response can be read by json.loads().\\

User's Question: \{question\}\\
Table File Path: \{table\_path\}\\
Table:\\
\{csv\_data\}\\

Response: \\
\end{tcolorbox}

\begin{tcolorbox}[
  colback=white, colframe=gray!80,
  title=Prompt of Table Describer, fonttitle=\bfseries,
  boxrule=0.8pt, arc=4pt,
  left=6pt, right=6pt, top=6pt, bottom=6pt,fontupper=\small
]
Given a database schema regarding \{table\_name\}, your task is to analyse all columns in the database and add detailed explanations for the database and each column.

\textbf{Requirements:}
\begin{enumerate}[leftmargin=*, noitemsep]
  \item Response should include column names and the specific meanings of each column to help users better understand the data content.
  \item Response format example:\\
\{\\
  \hspace*{2em}``table\_description": ``...", \\
  \hspace*{2em}``column\_description": [\\
  \hspace*{4em}{``column\_name": ``Age", ``specific\_meaning": ``Represents User's Age"},\\
\hspace*{4em}{``column\_name": ``Joined Date", ``specific\_meaning": ``The date user joined."},\\
\hspace*{4em}{``column\_name": ``Gender", ``specific\_meaning": ``User's Gender"},\\
\hspace*{4em}{``column\_name": ``City", ``specific\_meaning": ``City where the user resides"}\\
  \hspace*{2em}]\\
\}\\
  \item Definition of fields:
    \begin{itemize}
      \item Table\_Description: Explain the main content and possible uses of the table.
      \item Column\_Description: Explain the meaning of each column.
    \end{itemize}
  \item Ensure that the response format is a compact and valid JSON format without any additional explanations, escape characters, line breaks, or backslashes.
\end{enumerate}

\textbf{Database Schema}\\
\{table\_schema\}

Please respond in \textbf{JSON} format complying with the above requirements.
\end{tcolorbox}

\vspace{1.5em}

\begin{tcolorbox}[
  colback=white, colframe=gray!80,
  title=TableZoomer Prompt for Query Router, fonttitle=\bfseries,
  boxrule=0.8pt, arc=4pt,
  left=6pt, right=6pt, top=6pt, bottom=6pt,
  breakable ,fontupper=\small
]
As an experienced and professional data analysis assistant, your goal is to analyze a user's question and identify the necessary data from the table schema to help answer the query. The table schema consists of a table description and multiple column descriptions.\\

\textbf{Task}
You need to produce a JSON object with the following structure:

[\\
\hspace*{2em}  \{\\
\hspace*{4em}    ``type": ``row-column retrieval / column-only retrieval",\\
\hspace*{4em}    ``relevant\_column\_list": [``col1", ``col2", ...],\\
\hspace*{4em}    ``row\_match\_list": [{``entity name": ``column name"}, ...]\\
\hspace*{2em}\}\\
]\\

\textbf{Explanation of Fields}

``type": Choose one of the two options: \\

 - ``column-only retrieval": The question can be answered by analyzing or aggregating over some columns without requiring specific row filtering. \\

 - ``row-column retrieval": The question requires filtering specific rows based on values (entities) mentioned in the query and then analyzing columns. \\

``relevant\_column\_list": Identify all necessary columns from the table schema to answer the query. Always use column names exactly as provided. Use column descriptions and examples to disambiguate similar fields. When queries are ambiguous, include all potentially relevant columns. Never include columns not in the schema.\\ 

``row\_match\_list": Only extract when the query requires row-level filtering. Match entity values from the query to their corresponding column names. If not applicable, use an empty list []. \\

\textbf{Output Format}
Please strictly respond in the following JSON format: 

[\\
\hspace*{2em}\{\\
\hspace*{4em}``type": ``...",\\
\hspace*{4em}``relevant\_column\_list": [``..."],\\
\hspace*{4em}``row\_match\_list": [{``...": ``..."}, ...]\\
\hspace*{2em}\}\\
]

\textbf{Examples}

Question: What is the average concentration of PM2.5 in Sichuan Province in January 2015?  \\
Response:

[\\
\hspace*{2em}\{\\
\hspace*{4em}``type": ``row-column retrieval",\\
\hspace*{4em}``relevant\_column\_list": [``date", ``province", ``PM2.5"],\\
\hspace*{4em}``row\_match\_list": [\{``Sichuan Province": "province"\}, \{``January 2015": ``date"\}]\\
\hspace*{2em}\}\\
]

Question: What is the average fare for all passengers by class?  \\
Response:

[\\
\hspace*{2em}\{\\
\hspace*{4em}``type": ``column-only retrieval",\\
\hspace*{4em}``relevant\_column\_list": [``fare", ``class"],\\
\hspace*{4em}``row\_match\_list": []\\
\hspace*{2em}\}\\
]\\

\vspace{0.5em}
\textbf{Now let's begin!}

\textbf{Table Schema}\\
\{table\_schema\}

\textbf{Question:} \{query\} \\

\textbf{Response:}
\end{tcolorbox}

\vspace{1.5em}

\begin{tcolorbox}[
  colback=white, colframe=gray!80,
  title=Prompt of ReAct, fonttitle=\bfseries,
  boxrule=0.8pt, arc=4pt,
  left=6pt, right=6pt, top=6pt, bottom=6pt,,fontupper=\small
]
As an intelligent assistant for table analysis, your primary task is to analyze the table schema and assist in answering questions based on the data. To perform this, follow these guidelines:
\begin{enumerate}[leftmargin=*, noitemsep]
  \item You cannot view the table directly. However, you are provided with schema details and some sample cell values.
  \item Use these schema details to frame relevant Python queries that progressively solve the user's question.
  \item Strictly adhere to the structured format below to document your thought process, actions, observations, and responses.
\end{enumerate}

\textbf{Provided Information:}  \\
Schema Retrieval Results:\\
\{table\_schema\}\\

\textbf{Thinking Format:}
\begin{itemize}
  \item Query: Input question that needs to be answered.
  \item Thought: Clearly state your reasoning and plan of action.
  \item Action: Generate concrete Python-based ideas based on the schema to retrieve observations or answers.
  \item Observation: Record the result or insight from the action. If not available, note the ambiguity or missing info.
  \item Repeat Thought → Action → Observation as needed.
  \item Thought: After gathering enough observations, decide if the question can be answered and summarize findings.
  \item Response: Present a concise and accurate answer to the original question.
\end{itemize}

\textbf{Task:}  
Given the table schema retrieval results above, analyze the input question and generate the thought process or final response in the structured format.

\textbf{Input Question:} \\ 
\{query\}

\textbf{Thinking Process Records:}

\{his\_observations\}

\end{tcolorbox}

\vspace{1.5em}

\begin{tcolorbox}[
  colback=white, colframe=gray!80,
  title=Prompt of Answer Formatter, fonttitle=\bfseries,
  boxrule=0.8pt, arc=4pt,
  left=6pt, right=6pt, top=6pt, bottom=6pt,,fontupper=\small
]
Based on the following table schema and thought process records, generate the final brief answer of the user query \{query\} to the table.

\textbf{Rules:}
\begin{enumerate}[leftmargin=*, noitemsep]
  \item Thoroughly analyze the connection between the query and the thought process, and extract the correct answer.
  \item Determine the data type of answer based on the understanding of the user question. The data type of final answer must be one of the following:\\
\{requirements\} \\
  \item Output the answer directly without any prefix words or explanations.
  \item The final answer value must be derived from the values extracted from the provided data. Do not perform rewriting, expansion, or format conversion.Prohibit changing the type of source data.
\end{enumerate}

\textbf{Table Schema}\\
\{table\_schema\}

\textbf{Here are your thought process records}

\{thought\_process\}

\textbf{User Query}  \\
Query: \{query\}

\textbf{Final Answer:}
\end{tcolorbox}

\vspace{1.5em}




\end{appendices}


\bibliography{main}

\end{document}